\documentclass[10pt,twocolumn,letterpaper]{article}

\usepackage{cvpr}              

\usepackage{graphicx}
\usepackage{amsmath}
\usepackage{amssymb}
\usepackage{booktabs}
\usepackage{stfloats}
\usepackage{cite}
\usepackage{color}
\usepackage{times}
\usepackage{overpic}
\usepackage{bm}
\usepackage{tabu}
\usepackage{multicol}
\usepackage{multirow}
\usepackage{scalerel,stackengine}
\usepackage[table ]{xcolor}
\usepackage[accsupp]{axessibility}
\newcommand{\PAR}[1]{\vskip3pt \noindent{\bf #1~}}
\usepackage[pagebackref,breaklinks,colorlinks]{hyperref}

\usepackage[capitalize]{cleveref}
\crefname{section}{Sec.}{Secs.}
\Crefname{section}{Section}{Sections}
\Crefname{table}{Table}{Tables}
\crefname{table}{Tab.}{Tabs.}

\def\cvprPaperID{3124} 
\def\confName{CVPR}
\def\confYear{2022}

\begin{document}

\title{HSC4D: Human-centered 4D Scene Capture in Large-scale Indoor-outdoor Space Using Wearable IMUs and LiDAR}

\author{Yudi~Dai\textsuperscript{1} \hspace{2mm}Yitai~Lin\textsuperscript{1}\hspace{2mm}Chenglu~Wen\textsuperscript{1,}\thanks{Corresponding author.}\hspace{2mm}   Siqi~Shen\textsuperscript{1} \hspace{2mm}Lan~Xu\textsuperscript{2}\hspace{2mm}Jingyi~Yu\textsuperscript{2}\hspace{2mm}Yuexin~Ma\textsuperscript{2}\hspace{2mm} Cheng~Wang\textsuperscript{1}\\
$^{1}$Xiamen University, China \hspace{2mm} $^{2}$ShanghaiTech University, China}

\makeatletter
\let\@oldmaketitle\@maketitle
\renewcommand{\@maketitle}{
   \@oldmaketitle
	\begin{center}
      \vspace{-6mm}
      \includegraphics[width=0.96\linewidth]{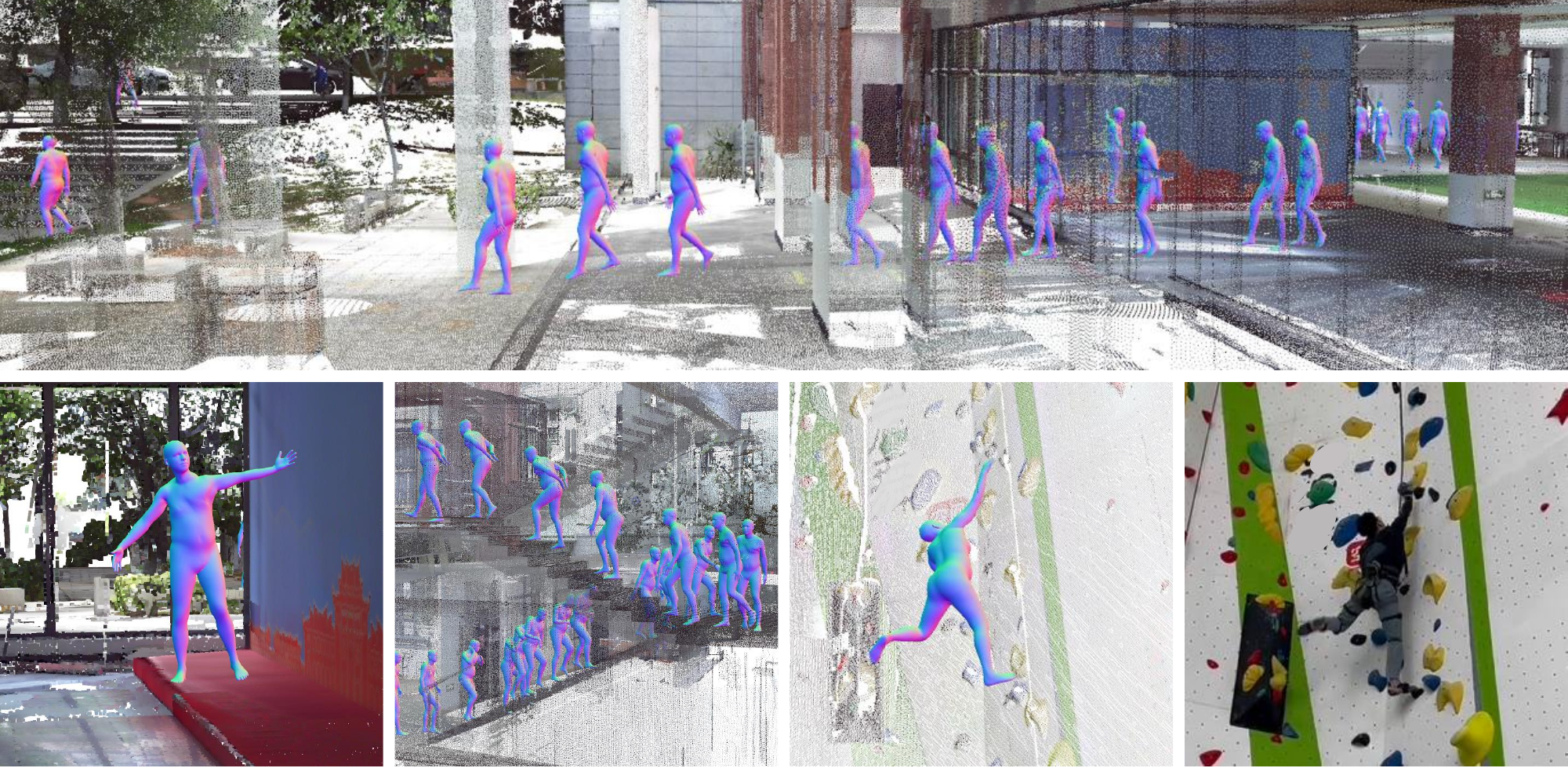}
	\end{center}
   \vspace{-2mm}

  \refstepcounter{figure}\normalfont Figure~\thefigure. 3D human motions with accurate localization in diverse, challenging scenes. The top line shows a person walking from indoor to outdoor. The bottom row figures show the challenge cases, where the bottom right figure is the rock climbing's third-view from the camera. 
	\label{fig:overview}
	\newline
   }
\makeatother

\maketitle
\thispagestyle{empty}
\enlargethispage{\footskip}
      
\begin{abstract}
   We propose Human-centered 4D Scene Capture (HSC4D) to accurately and efficiently create a dynamic digital world, containing large-scale indoor-outdoor scenes, diverse human motions, and rich interactions between humans and environments. 
   Using only body-mounted IMUs and LiDAR, HSC4D is space-free without any external devices' constraints and map-free without pre-built maps. Considering that IMUs can capture human poses but always drift for long-period use, while LiDAR is stable for global localization but rough for local positions and orientations, HSC4D makes both sensors complement each other by a joint optimization and achieves promising results for long-term capture. Relationships between humans and environments are also explored to make their interaction more realistic. 
   To facilitate many down-stream tasks, like AR, VR, robots, autonomous driving, etc., we propose a dataset containing three large scenes (1k-5k $m^2$) with accurate dynamic human motions and locations. Diverse scenarios (climbing gym, multi-story building, slope, etc.) and challenging human activities (exercising, walking up/down stairs, climbing, etc.) demonstrate the effectiveness and the generalization ability of HSC4D. The dataset and code is available at \url{http://www.lidarhumanmotion.net/hsc4d/}.  
\end{abstract}
\vspace{-2ex}

\section{Introduction}
\label{sec:intro}

The development of digital society is overwhelming because it can enrich peoples' life by serving Augmented Reality, Virtual Reality, smart city, robots, autonomous driving, etc. Humans and environments are two main components for creating the digital world. Current research works always separate dynamic human motions and static environments. Actually, taking account their interactions can help improve both capture accuracy. It is a trend to directly capture the whole scene with consecutive human activities. 

To capture human motions, IMU sensors are widely used and always be mounted on different parts of the human body, like arms, legs, feet, head, etc. It can capture accurate short-term motions but suffer from severe drift with the acquisition time increasing. Some methods~\cite{Marcard_2018_ECCV, kaichi2020resolving, xu2019flyfusion, xu2017flycap, dou2016fusion4d} utilize extra external RGB or RGBD cameras as a remedy to improve the accuracy, but result in limited capture space, human activities, and interactions. HPS~\cite{guzov2021human} uses a head-mounted camera, which looks outwards like the human eyes, to complement IMUs in global localization. Without the constraint of external cameras, it can recover the full-body pose and register the human in large 3D scan of real scenes. However, HPS requires pre-built maps and a huge image database for self-localization. 

For accurate localization and mapping~\cite{zhang2014loam}, LiDAR is the most applicable sensor in current days, which is popular for mobile robots and autonomous vehicles. LiDAR is also extensively used for large-scale scene scans. Although there are many LiDAR-captured datasets, including indoor scenes~\cite{caesar2020nuscenes, romero2017inlida} and large-scale outdoor scenes~\cite{geiger2013vision, maddern20171}, they focus on scene understanding and 3D perception, ignoring accurate human poses. PedX~\cite{kim2019pedx} provides 3D poses of pedestrians by using SMPL~\cite{smpl2015loper} parameterization for joint locations of instances on third-person-view images, which is not accurate as IMUs. Furthermore, it focuses on traffic scenes and is not applicable for generating diverse 3D human motions.

 Taking advantage of IMUs-based motion caption and LiDAR-based localization and scene capture, we propose Human-centered 4D Scene Capture (HSC4D) to accurately and efficiently create a dynamic digital world with consecutive human motions in indoor-outdoor scenes. Using only body-mounted sensors, HSC4D is space-free and pose-free, and the interaction between humans and the environment inside is also free, which makes it possible to capture most of the human-involved real-world scenes. Compared with camera-based localization, LiDAR is more precise for global localization, which dramatically reduces the drift of IMUs, and does not need pre-built maps. IMUs can improve the accuracy of LiDAR-captured local trajectories, where the error is caused by the jitter of the body. Making use of the complement of both sensors, we propose a joint optimization to improve the performance of motion estimation and human-scene mapping by considering several physical constraints.
 
To facilitate further research and down-stream applications, we propose a dataset containing three large scenes (1k-5k $m^2$) with accurate dynamic human motions and locations. As Fig.~\ref{fig:overview} shows, the dataset contains diverse scenarios, like climbing gym, multi-story building, slope, etc., and challenging human activities, such as exercising, walking up/down stairs, climbing, etc. Accurate human poses and natural interactions between human and the environment demonstrate the effectiveness and the generalization ability of HSC4D.

Our contributions are summarized as follows: 
\vspace{-2mm}
\begin{itemize}
\item Based on body-mounted IMUs and LiDAR, we propose Human-centered 4D Scene Capture (HSC4D) for creating a human-centered dynamic digital world, which is space-free, pose-free, and interaction-free.
\vspace{-2mm}

\item We propose a joint optimization method by integrating LiDAR SLAM results and IMU poses with scene constraints, resulting in natural human motions and accurate global localization in large scenes. 
\vspace{-2mm}

\item We provide a new dataset containing LiDAR point cloud of large-scale scenes, IMU data of human poses, and the results of poses and mapping by our optimization, which also demonstrates the effectiveness and the generalization ability of HSC4D. 
\end{itemize}

\section{Related work}
\label{sec:Related work}
\subsection{IMU sensors for human pose estimation}
IMU sensors have been widely used to capture human motions~\cite{roetenberg2007moven, vlasic2007practical, SIP,DIP:SIGGRAPHAsia:2018}. However, IMU-based methods suffer from severe drift over time. To improve the pose estimation accuracy, some methods~\cite{Marcard_2018_ECCV, kaichi2020resolving, dou2016fusion4d, Marcard2016HumanPE, Malleson3DV17} utilize extra external RGB or RGBD cameras as a remedy. 
Helten \etal~\cite{Helten:2013} combined two RGB-D cameras with IMUs to perform local pose optimization.
HybridFusion~\cite{zherong2018} has achieved more accurate motion tracking performance by combining an RGBD camera with multiple IMUs.
3DPW~\cite{Marcard_2018_ECCV} uses a single hand-held RGB camera and IMUs to optimize human pose for a certain period of frames simultaneously.
Constraints from external cameras assist in recovering more accurate 3D poses but result in limited capture space, human activities, and interactions. 
HPS~\cite{guzov2021human} uses a first-view head-mounted camera to self-localize the 3D pose from IMUs to the scene. 
However, HPS requires pre-built maps and an image database for self-localization.  Instead, we use only body-mounted IMUs and LiDAR. Without any external devices' constraints and any pre-built maps, we achieve promising results for long-term human motion capture.

\subsection{Human self-localization methods}
Human self-localization aims at estimating the 6-DoF of the human subject with carrying devices. The received signal strength (RSS) fingerprinting-based methodologies~\cite{abbas2019wideep, lemic2014infrastructure, alarifi2016ultra} are widely used for indoor human localization. However, these methods need external receivers and are limited to the indoor space. Some image-based methods~\cite{kendall2015posenet, radwan2018vlocnet++, wang2020atloc} regress locations directly from a single image with a pre-built map. Still, the scene-specific property makes them hard to generalize to unseen scenes. Some methods integrate IMU as an aid sensor \cite{shan2020lio, oleynikova2015real} to improve accuracy. With robustness and low drift, LiDAR-based localization has been successfully applied in indoor \cite{wang2018single, peng2017lidar} and outdoor \cite{yin2018locnet, uy2018pointnetvlad, yu2021deep, li2019net} scenes. To localize the human subject, LiDAR are designed as backpacked \cite{Liu2010IndoorLA, 8736839, Karam2019DesignCA} and hand-held\cite{bauwens2016forest}. LiDAR-based localization systems are usually big pieces of equipment and would affect human motion. We design a lightweight hip-mounted LiDAR to rigidly connect with the human body, achieving human self-localization in both large indoor and outdoor scenes.

\subsection{LiDAR-based mapping methods}
LiDAR is currently the most applicable sensor for 3D mapping. As a pioneer, zhang \etal \cite{zhang2014loam} proposed LOAM, a real-time odometry and mapping method using a LiDAR, greatly boosting the 3D mapping research. Some methods ~\cite{shan2018lego, wang2021, lin2020loam, jiao2021robust} further improve LOAM mapping for specific scenes and sensors. LeGo-LOAM\cite{shan2018lego} is a ground-optimized version, which requires to keep the LiDAR horizontal. LiDAR-based methods tend to fail when the $Z$-axis jitters severely. To address the drift problem and improve robustness, more sensors, such as visual sensors\cite{zhang2015visual, shin2020dvl, seo2019tight}, IMU\cite{shan2020lio, geneva2018lips, opromolla2016lidar}, or both\cite{deilamsalehy2016sensor, zuo2019lic, shan2021lvi}, have been integrated in mapping task.
To make the system lighter and able to work wirelessly, we propose a LiDAR-only method for localization and mapping in the scene. The joint optimization result with scene and IMU poses will further improve the LiDAR mapping result. 

\section{System setup}
\label{sec:system setup}
\subsection{Notations and Task Description}
The problem addressed in this paper is to estimate the 3D human motion with a 3D spinning LiDAR and IMUs in a large unknown scene and build a map for it, where human motion includes local 3D pose and global localization.

\PAR{Notations.} The $N$ frames human motion is represented as $M=(T, \theta, \beta)$, where $T$ is the $N\times3$ translation parameter, $\theta$ is the $N\times24\times3$ pose parameter, and $\beta$ is the $N\times10$ shape parameter. We assume that $\beta$ is constant during a data recording. The 3D point cloud scene is represented as \bm{$S$}. We use right subscript $k, k\in~Z^+$ to indicate the index of a frame.
We use the Skinned Multi-Person Linear (SMPL) body model \cite{smpl2015loper} $\varPhi(\cdot)$ to map $M_k$ to human mesh models $V_k, V_k\in~\mathbb{R}^{6890\times3}$.

Let us define three coordinate systems: 1) IMU coordinate system \{$I$\}: origin is at the hip joint of the first SMPL model, and $X/Y/Z$ axis is pointing to the right/upward/forward of the human. 2) LiDAR Coordinate system \{$L$\}: origin is at the center of the LiDAR, and $X/Y/Z$ axis is pointing to the right/forward/upward of the LiDAR. 3) Global coordinate system \{$W$\}: the first LiDAR frame's coordinate.

\PAR{Task definition.} Given a sequence of LiDAR sweep $P_k^L, k\in~Z^+$ in \{$L$\} and a sequence of 3D Human motion $\bm{M}_k^I$ in \{$I$\}, compute the human motion $\bm{M}_k^W$ in \{$W$\} and build the 3D scene $\bm{S}$ with $P_k^I$. 

\subsection{System design}
\begin{figure}[!htb]
    \centering
     \includegraphics[width=0.85\linewidth]{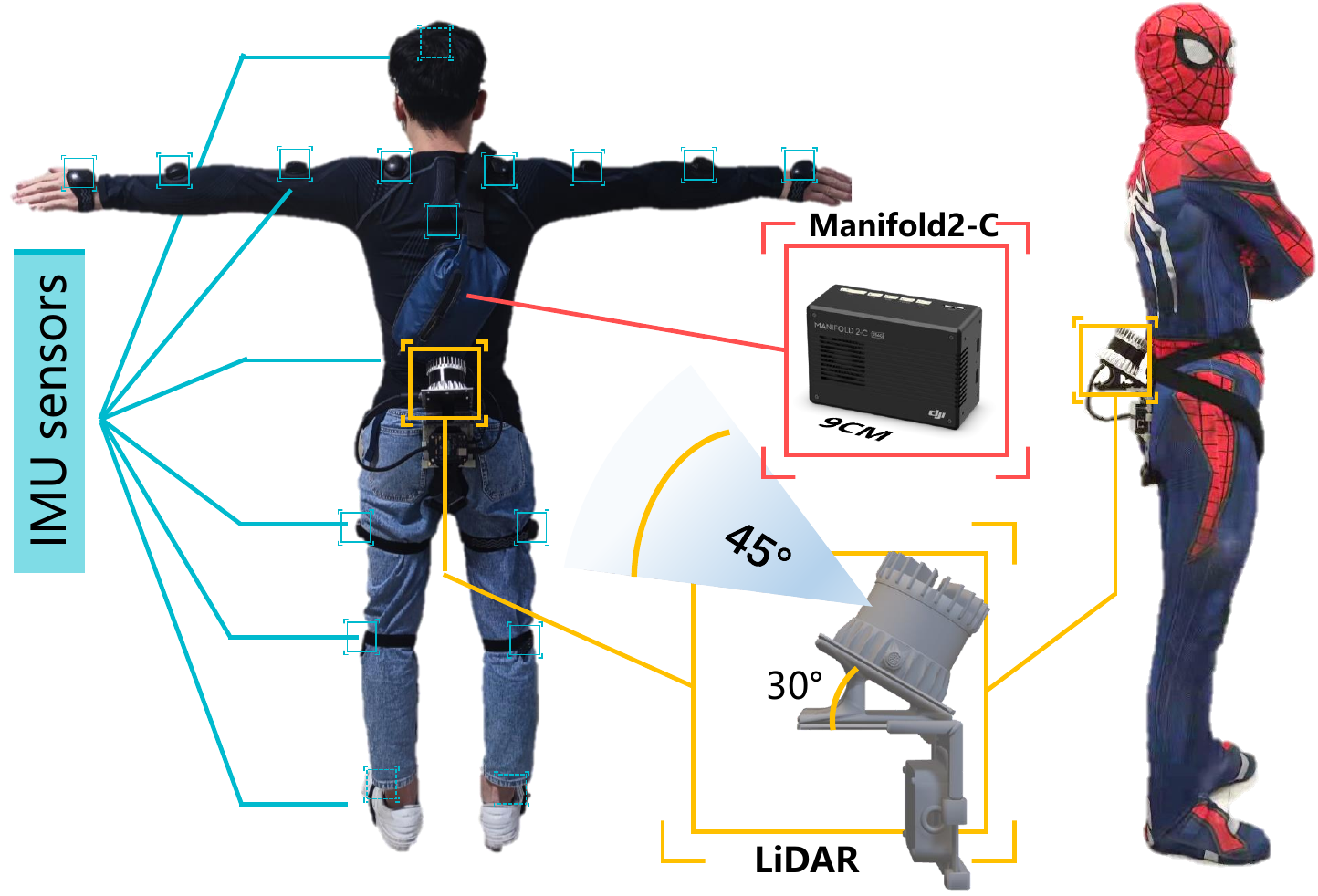}
     \vspace{-1mm}
     \caption{Overview of our capturing system, which includes 17 body-attached IMU sensors, a LiDAR sensor, and a mini-computer.}
     \label{fig:system_design}
    \vspace{-4mm}
\end{figure}

\PAR{Hardware.} We use a 64-beams Ouster LiDAR to acquire 3D point clouds $P_k^L$, and Noitom's inertial MoCap product PN Studio to obtain human motion $M_k^I$. The PN Studio uses 17 IMUs attached to the body limbs and a wireless receiver to acquire data. To make the LiDAR work wirelessly, we connect the LiDAR to a $9cm\times6cm\times3cm$ DJI Manifold2-C mini-computer and use a 24V mobile power to charge the LiDAR and the computer. To ensure a lightweight and precise capturing system, We modified all cables and designed a $10cm\times10cm$ L-shape bracket to mount the LiDAR package. The battery and Manifold2 are stored in a small bag on the human's back. The LiDAR is worn tightly close to the hip bone, making the origins of \{$I$\} and \{$L$\} as close as possible. Thus, we assume that LiDAR and IMUs have a rigid transformation.

The  Ouster LiDAR has a 360° horizon view and a 45° vertical view. However, due to the occlusion caused by back and swing arms, the horizon field of view is reduced, ranging from 150° to 200°. To avoid most laser points hitting the nearby ground, we tilt up the LiDAR for 30° to get a good vertical scanning view.

\section{Approach}
\label{sec:method}
\begin{figure*}[!htb]
    \centering
     \includegraphics[width=0.95\linewidth]{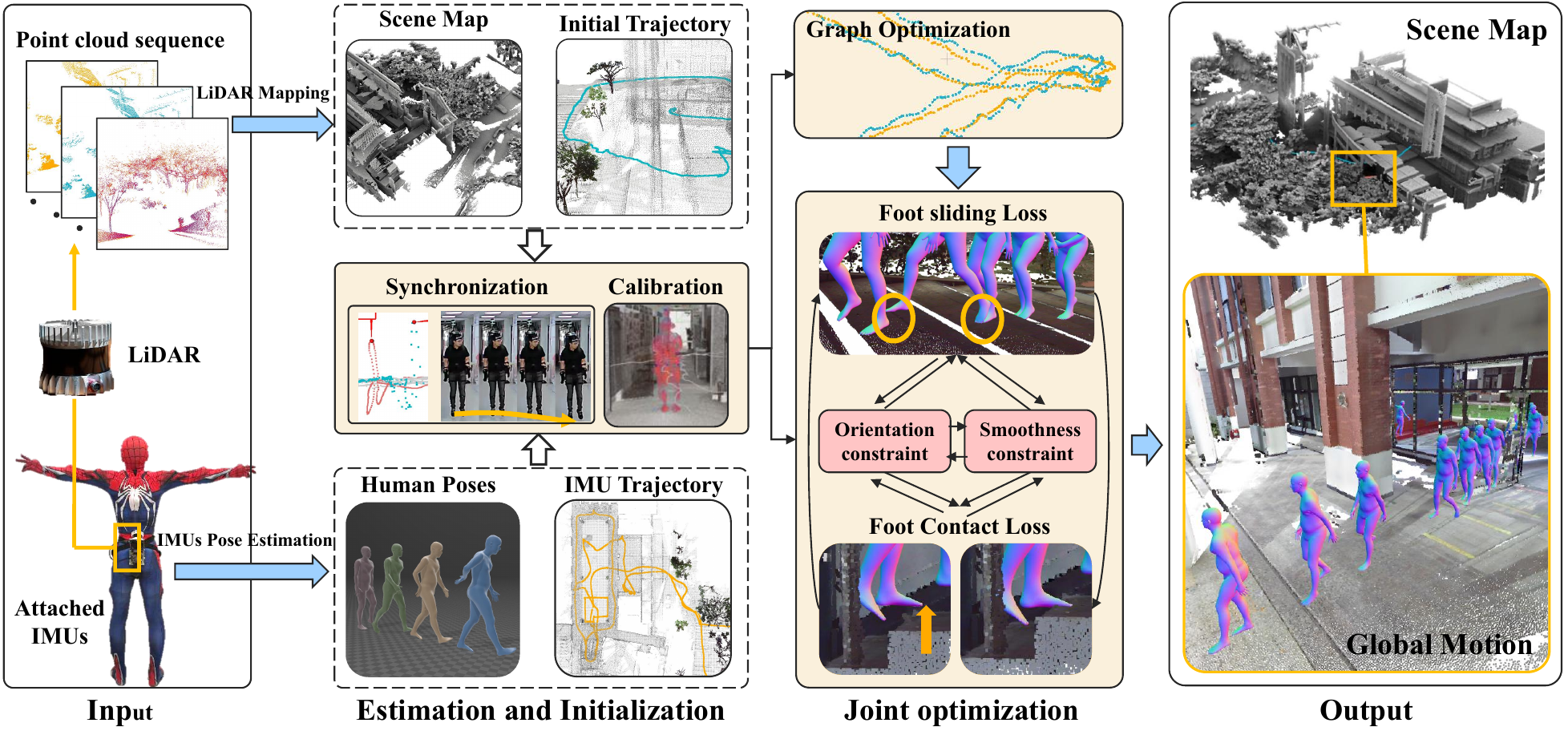}
     \vspace{-1mm}
  
     \caption{\textbf{Overview of our method. }The inputs are point cloud sequence and IMUs data. IMUs pose estimation and LiDAR mapping are performed, respectively. Synchronization and calibration are applied to prepare data for further optimization. Finally, graph-based optimization and joint optimization are performed to produce the global motion in the scene map.  }
     \label{fig:method}
    \vspace{-3mm}
\end{figure*}

We first obtain the 3D human motion output by the inertial MoCap system. Second, we estimate the ego-motion of LiDAR and build a 3D scene map \bm{${S} $} through point cloud data $P_k^I$. Then we perform a data initialization to prepare data for further optimization. Later we perform a graph-based optimization to fuse LiDAR trajectory and IMU trajectory. Finally, by combining the LiDAR data, IMUs data, and 3D scene, a joint optimization is performed to give the human motion $\bm{M}$ and an optimized scene.

\subsection{IMUs Pose Estimation}
This subsection aims to estimate the motion $\bm{M^I}=(T^I, \theta^I, \beta)$ in IMU coordinate \{$I$\}, where $T^I$ and $\theta^I$ are provided by the commercial MoCap product. Pose parameter $\theta^I$ is composed of the hip joint's orientation $R^I$ relative to the start frame and other joints' rotation relative to their parent joint. $T_k^I$ indicates the $k$-th frame translation relative to the start. Since IMU is accurate in a short period, the relative value of $T^I$ and $R^I$ can be used in the optimization. 

\subsection{LiDAR Localization and Mapping}
Building a map using LiDAR data is challenging in this scene because the LiDAR jitters as the human walking and human body occludes the field of scanning view. By employing LiDAR-based SLAM methods\cite{zhang2014loam, wen2019toward}, we estimate the ego-motion of LiDAR and build the 3D scene map $\bm{S}$ with $P_k^L, k\in~Z^+$ in \{$L$\}. We first exact planer and edge feature points in every LiDAR scan $P_k^L$ and keep updating the feature map. Similar to \cite{wen2019toward}, we skip frame to frame odometry and only perform frame to map registration because the mapping process can run offline. Finally, Lidar's ego-motion $T^W$ and $R^W$, and the scene map $S$ are computed. The mapping function is denoted as:
\begin{equation}
    T^W, R^W, \bm{S} = \mathcal{F} (P_{1:N}^L) 
\end{equation} 

\subsection{Optimization initialization}
\label{sec:initailize}
\PAR{Coordinate calibration.} To obtain the rigid offset from LiDAR to IMU and make all coordinate systems aligned, we perform following steps: First, the human stands as an A-pose before capture, and the human's face direction is regarded as scene's $Y$-axis direction. After capturing, we rotate the scene cloud $Z$-axis perpendicular to the starting position's ground. Last, we translate the scene to make its origin to the first SMPL model's origin on the ground. LiDAR's ego motion $T^W$ and $R^W$ are translated and rotated as the scene does. To now, LiDAR data are calibrated to \{$W$\}. The pitch, roll, and translation of IMU are calibrated to the world coordinate. The IMU's yaw will be further refined in \cref{subsec:data fusion}.

\PAR{Time synchronization.} Firstly, we ask the subject to jump at the starting place of every capture. Secondly, we automatically locate the peaks in both $T^W$ and $T^I$ based on their $z$ value. Then, we can synchronize the LiDAR and IMU according to the two peaks' timestamps. Finally, we resample the IMU data (100Hz) to the same frame rate as LiDAR (20Hz).

\begin{figure}
    \centering
    \includegraphics[width=.92\columnwidth]{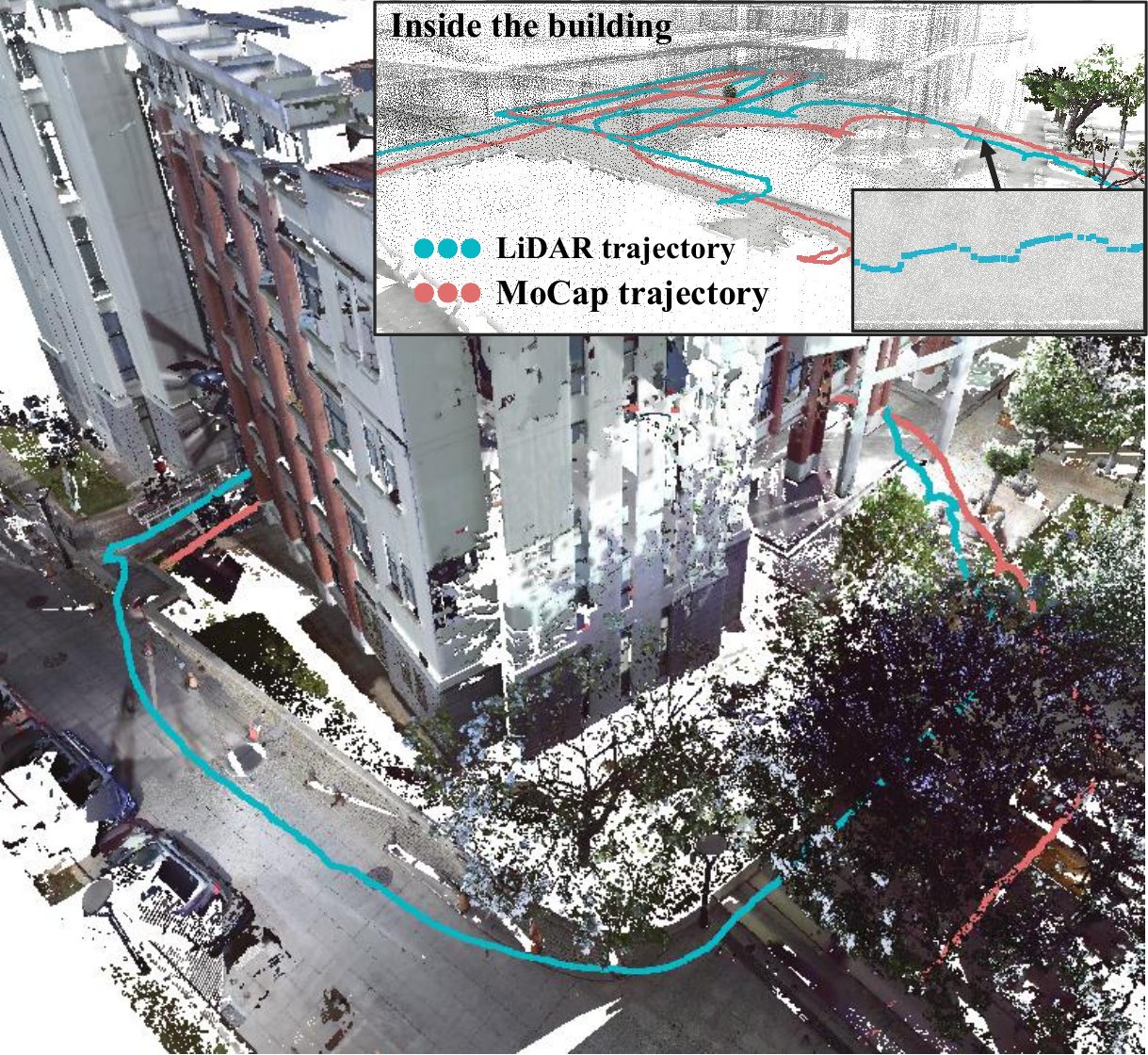}
   \vspace{-1mm}
   \caption{Comparison of trajectories of IMU and LiDAR.}
   \label{fig:traj_compare}
   \vspace{-2mm}
\end{figure}

\subsection{Graph optimization data fusion}
\label{subsec:data fusion}
As seen from \cref{fig:traj_compare}, IMUs drift severely over time and fail when the scene's height change, while the LiDAR localizes correctly but jitters at local movement. To estimate a more smooth and stable trajectory, we utilize both data's advantages. Our strategy is presented as follows: 1) first mark the frame in $T^W$ that exceeds $x$ (1.2$\sim$2.0) times of IMU velocity and the local fitted value as the outliers, 2) treat the remained $(R^W, T^W)$ as landmarks, and then segment $T^I$ every five seconds, 3) The nodes (IMU or landmark poses) and edges (a relative transformation between two nodes) construct a graph, 4) finally perform a graph optimization \cite{kummerle2011g2o} method to couple $T^I$ and $T^W$.

\subsection{Joint optimization}
\label{sec:optimization}
To obtain accurate and scene-natural human motion $\bm{M}=(T, \theta)$, and a higher quality scene cloud \bm{${S}$}, we perform a joint optimization method by using scene \bm{${S}$} and physics constraints. Then we send $T$ back to mapping function $\mathcal{F}$ as an initial value to create a new scene $\bm{S}_{opt}$. We use the following four constraints: the foot contact constraint $\mathcal{L}_{cont}$ encouraging the human standing on the ground, the sliding constraints $\mathcal{L}_{sld}$ eliminating the human walk sliding, the orientation constraint $\mathcal{L}_{ort}$ 
from $R^I$ making the rotation smooth, and the smoothness constraint $\mathcal{L}_{smt}$ making the translation smooth. The optimization is expressed as:
\begin{equation}
	\begin{split}
    &\mathcal{L}=\lambda_{cont}\mathcal{L}_{cont}+\lambda_{sld}\mathcal{L}_{sld}+{\lambda_{ort}\mathcal{L}}_{ort} 
        +{\lambda_{smt}\mathcal{L}}_{smt}\\
    &\bm{M}=arg\min_{\bm{M}}\mathcal{L}(\bm{M}|T^I, \theta^I,R^W, \bm{S})\\
    &\bm{S}_{opt}=\mathcal{F}(P_{1:N}^L, \bm{M}),
    \end{split}
\end{equation}
\noindent
where $\lambda_{cont}, \lambda_{sld}, \lambda_{ort}, \lambda_{smt}$ are coefficients of loss terms. $\mathcal{L}$ is minimized with a gradient descent algorithm to iteratively optimize $\bm{M}^{(i)}=(T^{(i)}, \theta^{(i)})$, where $(i)$ indicates the iteration. $\bm{M}^{(0)}$ is set as $(T^W, \theta^I)$,   

\PAR{Plane detection.} To improve validity of foot contact, we detect the planes near the human. We first use Cloth Simulation Filter (CSF) \cite{zhang2016ecsf} to extract ground points $\bm{S}_g$ in $\bm{S}$. And then we search neighboring points of $T^W$ in $\bm{S}_g$. Unlike the dense mesh model, the discrete point cloud has empty areas, resulting in invalid foot contact constraints. To adress this, we use RANSAC~\cite{schnabel2007efficient} to fit planes for the neighboring points. We denote the plane fuction as $p_k$.

\PAR{Foot contact constraint.} The foot contact loss is defined as the distance from a stable foot to its nearest ground. Unlike HPS knowing the information about which foot is stepping on the floor, we detect the foot state based on the movements. First, we compare the left and right foot movement for every successive foot vertices in ${V}_k^I=\varPhi(T_k^I,\theta_k^I,\beta)$ from IMU. One foot is marked as a stable foot if its movement is smaller than 2$cm$ and smaller than another foot's movement. The $k$-th frame's stable foot vertices index list in $V_j$ is denoted as $\mathcal{S}_{k}$ and the foot contact loss $\mathcal{L}_{cont}$ is written as:
\begin{equation}
	\begin{split}
        \mathcal{L}_{\text {cont}}=\frac{1}{l}\sum_{j=k}^{k+l}\sum_{v_{c} \in V_{j}^{\mathcal{S}_j}}
        \frac{1}{|\mathcal{S}_j|}\|v_{c}-\widetilde{v_{c}} p_j\|_{2},
    \end{split}
    \label{equ:cont}
\end{equation}
\noindent
where $\widetilde{v_{c}}$ is homogeneous coordinate of $v_c$. $V_{j}^{\mathcal{S}_j}$ is denoted as $\mathcal{S}_j$ foot vertices in $V_j=\varPhi(\bm{M}_j)$, which is from the motion to be optimized.

\PAR{Foot sliding constraint.} The foot sliding constraint reduces the motion's sliding on the ground, making the motion more natural and smooth. The sliding loss is defined as every two successive stable foot's distance:
\begin{equation}
	\begin{split}
        \mathcal{L}_{\text {slid}}=\frac{1}{l}\sum_{j=k}^{k+l-1} 
        \|\mathbb{E}{(V_{j+1}^{\mathcal{S}_{j+1}})}-\mathbb{E}(V_{j}^{\mathcal{S}_{j+1}})\|_{2},
    \end{split}
    \label{equ:sld}
\end{equation}
\noindent
where $\mathbb{E}(\cdot )$ is the average function.

\PAR{Orientation constraint.} This constraint encourages the motion $\bm{M}$ to rotate as smooth as IMU and have the same orientation with the landmarks $\mathcal{A}$ described in \cref{subsec:data fusion}. The orientation loss is written as follows: 
\begin{equation}
	\begin{split}
    &\mathcal{L}_{\text {orit }}=\frac{1}{|\mathcal{A}|}\sum_{j\in\mathcal{A}}\|(R_j)^{-1}R_j^W\|_{2} + \\
    &\frac{1}{l-1}\sum_{j=k}^{k+l-1}\max(0, \|(R_{j})^{-1}R_{j+1}\|_{2} - \|(R_{j}^I)^{-1}R_{j+1}^I\|_{2}).
    \end{split}
\end{equation}

\PAR{Smooth constraint.} This constraint encourages the human motion to move as smoothly as IMU motion, minimizing the difference between LiDAR and IMU sensors' translation distance. The smooth loss term is as follows:
\begin{equation}
	\begin{split}
	\mathcal{L}_{smt} = \frac{1}{l}\sum_{j=k}^{k+l-1}{\max(0, \|T_{k} - T_{k+1}\|_2 - \|T_{k+1}^I - T_{k}^I\|_2)}.
    \end{split}
\end{equation}

\section{Experiments}
\label{sec:experiments}

This section introduces our dataset and evaluates HSC4D in large indoor-outdoor 3D scenes. The results demonstrate the effectiveness and the generalization ability of HSC4D.  

\subsection{Dataset}
\label{sec:dataset}
\begin{figure*}[!htb]
    \centering
    \includegraphics[width=0.95\linewidth]{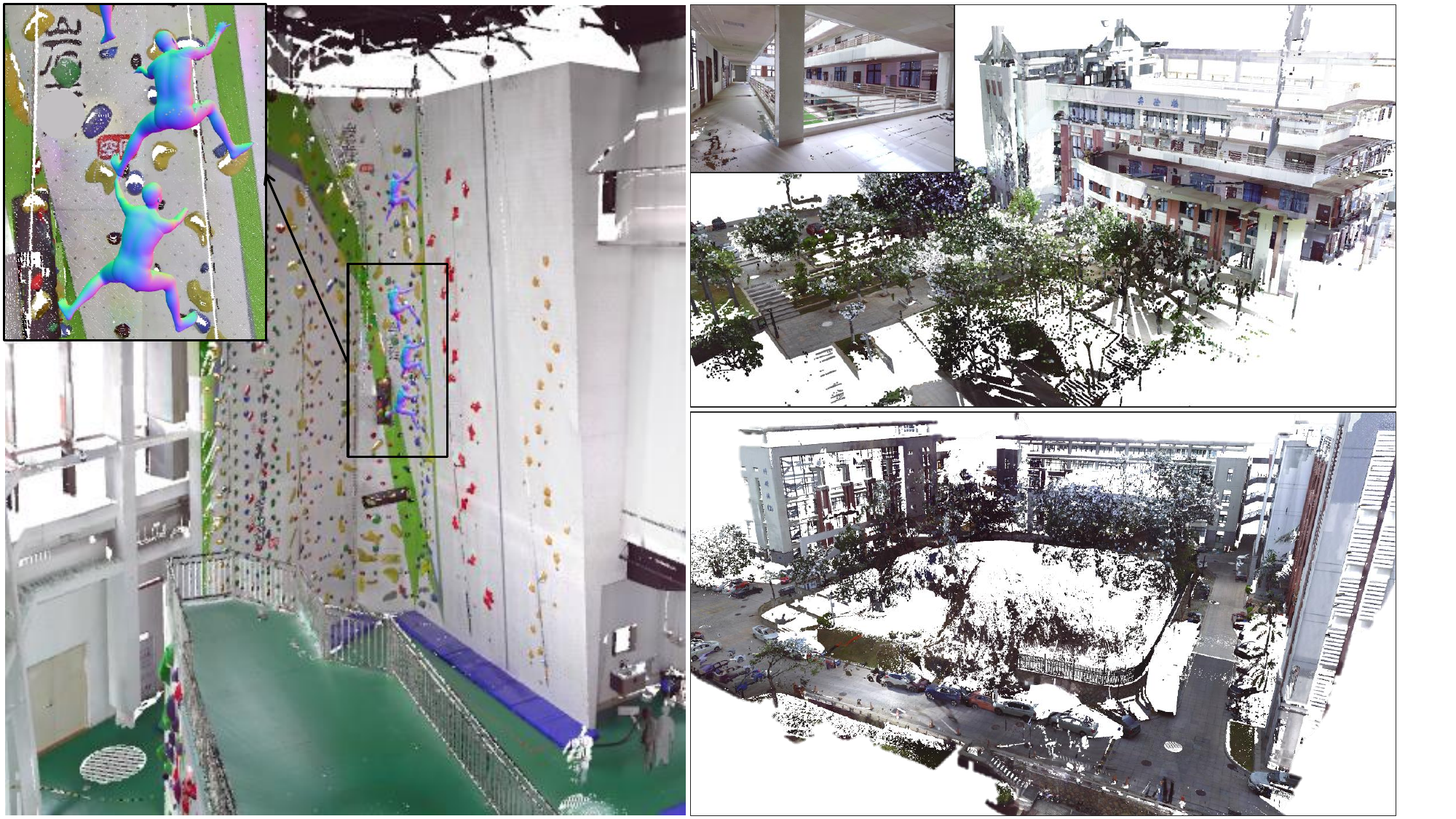}  
    \vspace{-1mm}
    \caption{The large indoor and outdoor scenes in our dataset. \textbf{Left}: a climbing gym (1200 $m^2$). \textbf{Right top}: a lab building with an outside courtyard 4000 $m^2$. \textbf{Right bottom}: a loop road scene 4600 $m^2$. 
    }
    \vspace{-2mm}
    \label{fig:dataset}
 \end{figure*}

 We propose an HSC4D dataset containing three large scenes: a rock climbing gym, a multi-story building, and an outdoor closed-loop road. The gym has a wall height of 20 meters, with a ground and climbing area size over 1200 $m^2$. The building scene's indoor and outdoor area size is up to 5000 $m^2$. The scenes have a diversity of heights and environments, including multi-story, slope, and staircase. The outdoor closed-loop road is $70m \times 65m$  with slope. In these scenes, captured human activities include walking, exercising, walking up/down stairs, rock climbing, speeching, etc. Since the 3D map from LiDAR lacks color information, we use a Terrestrial Laser Scanner (Trimble TX5) to scan the color scenes for better visualization. In summary, The HSC4D dataset provides 250K IMU frames (100Hz), 50k time-synchronized LiDAR frames (20Hz), our SLAM results, and colored ground truth point clouds of the scenes.

\vspace{-2mm}
\begin{table}[htb]
	\centering
    \footnotesize
	\centering
        \begin{tabular}{lll}
        \toprule
                         & \textbf{HPS} & \textbf{HSC4D} \\ 
         \midrule
                         & IMU-based Mocap & IMU-based Mocap \\ 
        \textbf{Sensors} & Head-mounted camera(*) & Hip-mounted LiDAR(*) \\ 
                         & \textbf{LiDARs with cameras(+)} &  \\ 
        \midrule
                        & IMU motions & IMU motions \\ 
         \textbf{Input} & Video frames(*) & LiDAR frames(*)\\ 
                        & \textbf{Pre-built 3D map(+)} & \\ 
        \bottomrule
        \end{tabular}
        \vspace{-1mm}
        \caption{\textbf{Comparison between HSC4D and HPS.}~~ (*): Different object. (+): Additional object. }
	\label{tab:comparison_data}
    \vspace{-5mm}
\end{table}
\subsection{Comparison}
The most related work to ours is HPS \cite{guzov2021human}, which uses IMUs, a head-mounted camera, and NavVis M6 equipped with 6 cameras and 4 LiDARs for human-scene modeling. 
Besides, HPS also heavily records all images registered to the captured 3D map for localization.
In contrast, We remove the tedious reliance on the pre-built map in HPS abtained using NavVis. Our approach is scene-prior-free and contributes a novel body-worn setting of LiDAR and IMUs for more practical human-scene modeling. (\cref{tab:comparison_data})

\noindent
\textbf{Baselines.} Since HSC4D is the first method in such scenes, there is no published baseline available to compare. For effective comparison, we name the IMU result as \textbf{{Baseline1}} and IMU + LiDAR without any optimization as \textbf{{Baseline2}}.

\noindent
\textbf{Global localization comparison.} First, in the ground truth point clouds provided by Trimble TX5, we mark some locations on the ground as checkpoints. Then the subject steps on the checkpoint during capturing. At last, we measure the distance from the SMPL model estimated by the method to the checkpoint as global localization error.

\noindent
\textbf{Local pose comparison.} To evaluate the local pose accuracy and smoothness, we compare the foot contact loss $\mathcal{L}_{cont}$ and the sliding loss $\mathcal{L}_{sld}$ described in \cref{sec:optimization}. The comparison is shown in \cref{tab:local_error}.

\begin{table*}[htb]
\centering
\footnotesize
\centering
\begin{tabular}{@{}cccccccccc@{}}
\toprule
\multirow{2}{*}{\textbf{Sequences}} & \multirow{2}{*}{\begin{tabular}[c]{@{}c@{}}$r$ = 0.6m \\ $l$ = 100\end{tabular}} & \multicolumn{3}{c}{\textbf{Loss term}} & \multicolumn{3}{c}{\textbf{Scene cropping radius}} & \multicolumn{2}{c}{\textbf{Length of optimization}} \\ \cmidrule(lr){3-5} \cmidrule(lr){6-8} \cmidrule(l){9-10} 
 & & w/o $\mathcal{L}_{cont}$ & w/o $\mathcal{L}_{sld}$ & w/o $\mathcal{L}_{smt}$ & $r = 0.4m$ & $r = 0.8m$ & $r = 1.0m$ & $l = 50$ & $l = 200$ \\ \midrule
Road  & 0.96/1.06 & +4.44/-0.03 & +0.01/+1.06 & -0.06/+0.43 & +0.02/+0.00 & -0.01/+0.00 & -0.01/+0.00 & -0.03/-0.06 & -0.02/-0.03 \\
    Gym01 & 0.66/0.65 & +7.03/-0.06 & +0.02/+0.70 & -0.31/-0.05 & +0.02/+0.00 & -0.01/+0.00 & -0.01/-0.01 & +0.00/+0.00 & +0.01/+0.00 \\
    Building01 & 0.80/1.20 & +5.47/-0.21 & +0.04/+0.77 & -0.10/-0.35 & +0.04/-0.12 & +0.02/-0.13 & +0.01/-0.13 & -0.03/-0.22 & +0.02/-0.13 \\
    Building04 & 0.92/0.96 & +5.88/-0.04 & -0.04/+0.86 & -0.12/-0.40  & +0.03/+0.00 & -0.01/+0.00 & -0.02/+0.01 & -0.02/+0.01 & -0.01/+0.00 \\
    \bottomrule
\end{tabular}
\vspace{-1mm}
\caption{\textbf{Quantitative evaluation of our method:} 
    We record $\mathcal{L}_{cont}$ and $\mathcal{L}_{sld}$ in column two. Column 3-5: loss term ablating comparisons. Column 6-8: analysis of neighborhood radius $r$ for cropping the scene. Column 9-10: analysis of the optimization sequence length $l$.}
\label{tab:quanti_eva}
\vspace{-4mm}

\end{table*}
\begin{table}[htb!]
	\centering
    \footnotesize
	\centering
    \begin{tabular}{ccccc}\toprule
        \textbf{CheckPoint} & \textbf{\begin{tabular}[c]{@{}c@{}}Distance \\ to start\end{tabular}} & \textbf{Baseline1} & \textbf{Balseline2} & \textbf{HSC4D} \\ \midrule
        Building $(P_1)$ & $ 35m $ & 65.96 & 8.14 & \textbf{7.01} \\ 
        Building$(P_2)$ & $ 350m $ & 214.74 & 17.79 & \textbf{11.02}  \\ 
        Building$(P_{3})$ & $ 480m $ & 331.84 & 24.45 & \textbf{18.11} \\ 
        Road$(P_{end})$ & $ 250m $ & 192.17 & 89.15 & \textbf{67.93} \\ \bottomrule
       \end{tabular}
    \vspace{-1mm}\caption{\textbf{Global localization error comparison between baselines and HSC4D}. \textbf{Baseline1}: IMU result. \textbf{Baseline2}: IMU pose + LiDAR localization. Error is measured as distance ($cm$) from the selected CheckPoint (\textit{CP}) to the SMPL model's foot.}
	\label{tab:compare_method}
	\vspace{-2mm}
\end{table}
\begin{table}[htb]
	\centering
    \footnotesize
	\centering
    \begin{tabular}{@{}ccccc@{}}
    \toprule
    \multirow{2}{*}{\textbf{Terrain}} & \multirow{2}{*}{\textbf{Sequences}} & \textbf{Baseline1} & \textbf{Baseline2} & \textbf{HSC4D} \\ \cmidrule(l){3-5} 
     &  & $L_{cont}$/$L_{sld}$ & $L_{cont}$/$L_{sld}$ & $L_{cont}$/$L_{sld}$ \\ \midrule
     Flat  & Building01 & 2.35/\textbf{1.00} & 6.22/2.57 & \textbf{0.80}/1.20 \\
     Flat  & Building03 & 3.54/\textbf{0.80} & 4.13/2.26 & \textbf{0.64}/0.91 \\
     Flat  & Gym02 & 4.20/0.43 & 5.33/1.08 & \textbf{0.78}/\textbf{0.20} \\
     Flat  & Gym03 & 44.156/1.14 & 3.21/2.08 & \textbf{0.79}/\textbf{0.26} \\
     Flat  & Gym01 & 187.65/\textbf{0.56} & 7.88/1.72 & \textbf{0.66}/0.65 \\
     Stairs & Building02 & 389.70/1.05 & 7.58/2.67 & \textbf{0.81}/\textbf{0.87} \\
     Stairs & Building04 & 394.46/\textbf{0.89} & 6.80/3.09 & \textbf{0.92}/0.96 \\
     Slope & Road  & 445.39/\textbf{1.03} & 8.59/3.70 & \textbf{0.96}/1.06 \\
 \bottomrule
    \end{tabular}
	\vspace{-1mm}
	\caption{\textbf{Local pose error comparison between baselines and HSC4D}. 
    \textit{Building01/03}: 1.5min/1min sequence on the second/first floor's corridor. \textit{Building02/04}: 3min/1min outdoor sequences, including stairs and slope. \textit{Gym01/02/03}: three one-minute walking and warming up sequences. \textit{Road}: 3.5 minute walking sequence.}
	\label{tab:local_error}
	\vspace{-3mm}
\end{table}

\begin{figure}[!htb]
    \centering
     \includegraphics[width=0.95\linewidth]{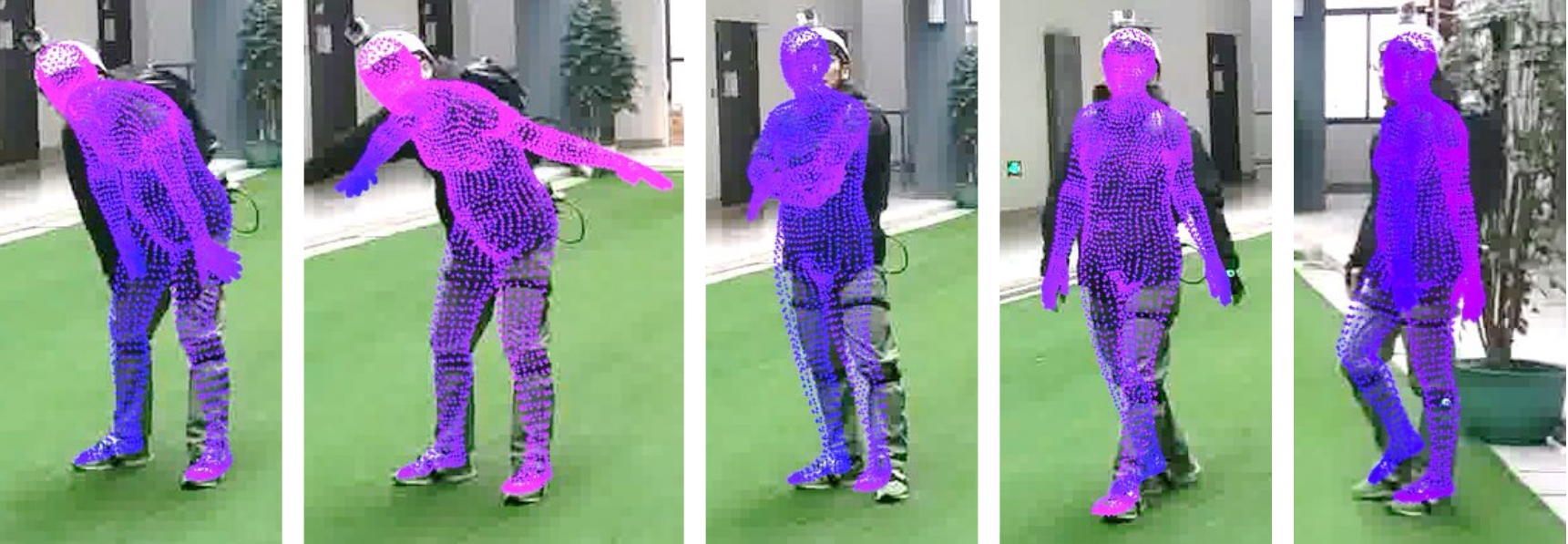}
     \vspace{-3mm}
     \caption{Qualitative results of local pose projected to images.}
     \label{fig:project}
    \vspace{-6mm}
\end{figure}

\begin{figure}[!htb]
    \centering
    \includegraphics[width=0.98\linewidth]{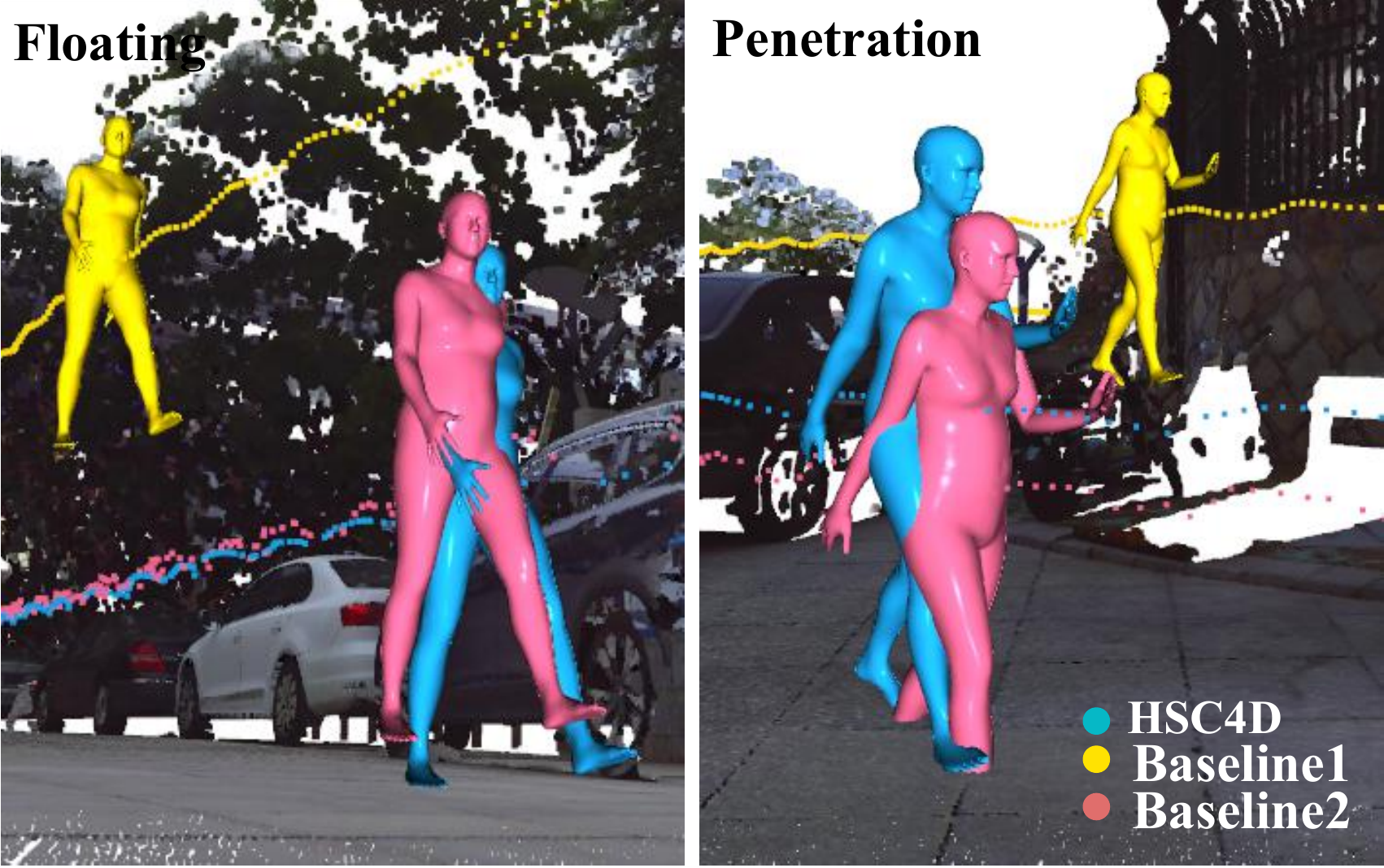}
    \caption{\textbf{An example of comparison between Baseline1, Baseline2, and HSC4D.} The colorful spheres represent the trajectories. HSC4D results are natural and accurate in both cases. Left: Baseline1 and Baseline2 are floating in the air. Right: Baseline1 is floating in the air, and Baseline2's lower leg penetrates the ground.}
    \vspace{-3mm}
    \label{fig:contact_loss}
    \vspace{-2mm}
 \end{figure}

\cref{tab:compare_method} shows the localization error comparisons between our method and baselines. As the distance increases in the scene, the error increases linearly in all methods. Baseline1's error is ten times compared to other methods because IMU drifts severely over time.
Baseline2 has a smaller global localization error, but its accumulative errors still vary from 8$cm$ to 90$cm$.
The last column shows that HSC4D achieves the smallest global localization errors in the multi-story building and the road with slopes. More specifically, HSC4D improves 78.3\% accuracy compared to Baseline1 and 25.4\% compared to Baseline2. In \cref{fig:contact_loss}, we show comparisons between baselines and our HSC4D.

\cref{tab:local_error} shows the comparison of local pose errors between our method and baselines. {Baseline1}'s foot contact loss is much larger than other methods especially in scenes where height change. Baseline2's $\mathcal{L}_{sld}$ is the largest among all methods. In the first three sequences where Baseline1 is not drift over height, \text{Baseline2}'s $\mathcal{L}_{cont}$ is much more larger than \text{Baseline1}. These cases indicate that LiDAR increases local errors. See from the last column, HSC4D significantly decreases $\mathcal{L}_{cont}$ in all cases and achieve comparable smoothness on $\mathcal{L}_{sld}$ compared to Baseline1. These comparisons reveal that HSC4D can achieve smooth results in local pose and is robust in various height diversity scenes.

\begin{figure*}[!htb]
    \centering
    \includegraphics[width=0.99\linewidth]{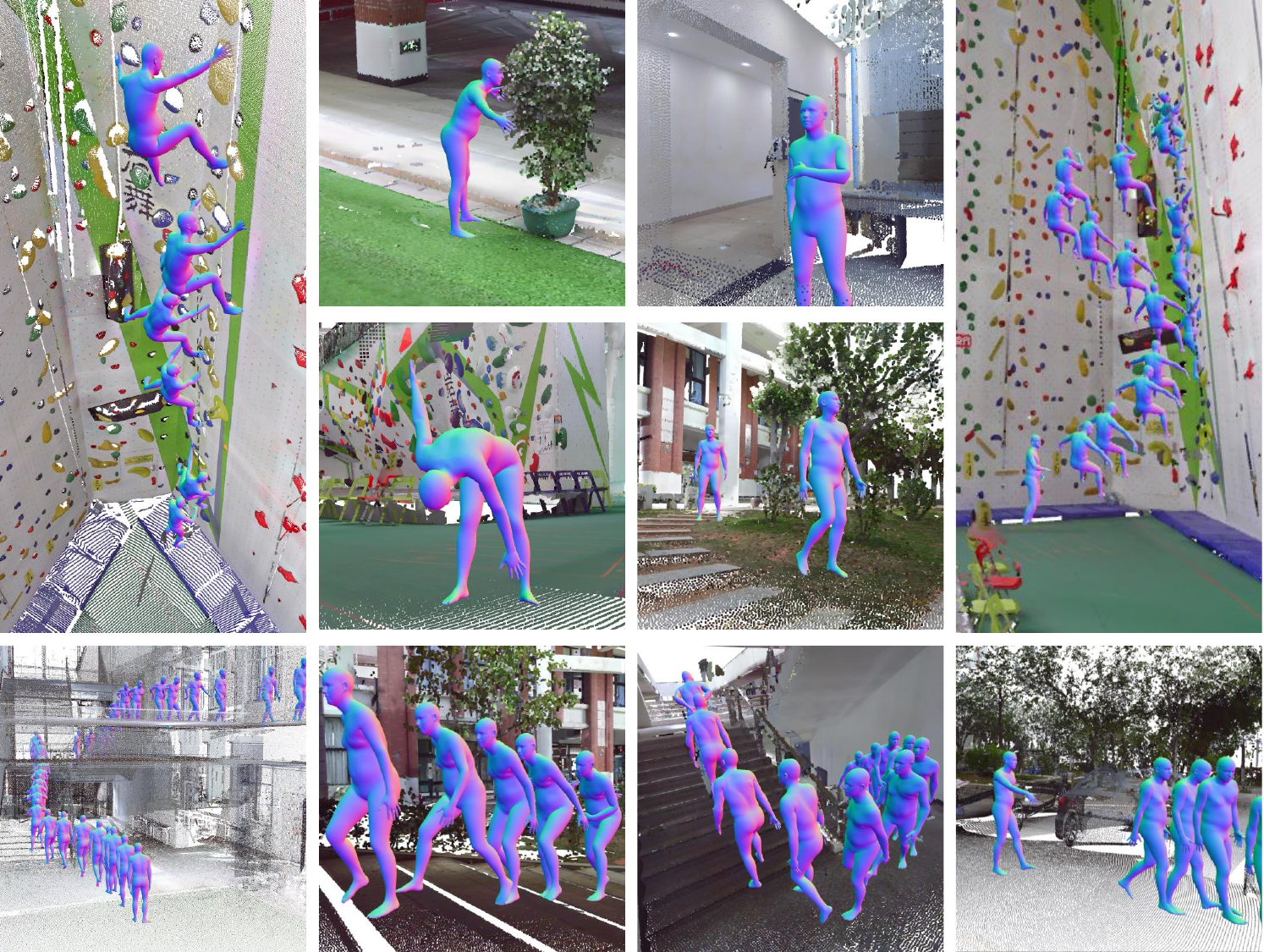} 
    \caption{Qualitative results of human activities in diverse scenes. The SMPL models in each figure are from the same human subject.}
    \vspace{-2mm}
    \label{fig:gallally}
    \vspace{-2mm}
 \end{figure*}

\subsection{Evaluation}

\PAR{Quantitative evaluation.}We evaluate our optimization method with ablating different loss terms by analyzing $\mathcal{L}_{cont}$ and $\mathcal{L}_{sld}$. We also analyze two parameters of the optimization: the neighborhood radius $r$, which is used to crop the scene for foot contact constraint, and the sub-sequence length $l$ of optimization. As shown in \cref{tab:quanti_eva}, without the foot contact constraint or the foot sliding constraint, $\mathcal{L}_{cont}$ or $\mathcal{L}_{sld}$ increases dramatically and has little impact on another term. Both $\mathcal{L}_{cont}$ and $\mathcal{L}_{sld}$ decrease without the smoothness constraint. However, the motion jitters more severely in 3D visualization. Overall, all loss terms are necessary to produce accurate and smooth human motion. We observe that both $r$ and $l$ have little impact on the result. In practice, to balance the computational resource and running time, we set $l=100$. And to ensure we can detect points in the scenes, we set $r=0.6m$.

\PAR{Qualitative evaluation.} For local pose qualitative evaluation, we use an extra camera registered to the scene and then project the human motions to the images \cref{fig:project}.
We show more qualitative examples in \cref{fig:gallally}. There are various human activities in different indoor and outdoor scenes shown in the figure. The sequence motions shown in figure represent the same human occurring at a successive time. Above examples show that our method can estimate the natural and challenging human motions with global localization. 

\section{Discussions}
\label{sec:discussion}

\PAR{Limitations.} First, HSC4D uses LiDAR SLAM results as the initial localization. Consequently, it is limited when LiDAR mapping fails, such as crowded places, narrow areas, etc. 
Second, to avoid a large occlusion during LiDAR data capturing, some activities are limited in our system, such as sitting in a chair with a backrest, standing back against a wall. 
Besides, the optimization loss terms in our pipeline are hand-created. Some cases like rock climbing do not always work. It is promising to propose a more general loss term or a deep learning framework cooperating with the semantic information. 

\PAR{Conclusions.} We present a Human-centered 4D Scene capture method to accurately and efficiently create a dynamic digital world using only body-mounted IMUs and LiDAR. Our method is space-free, pose-free, and map-free. By integrating LiDARs and IMUs, our proposed joint optimization algorithm can obtain accurate global localization and smooth local poses in large scenes. Additionally, we provide a new dataset containing large scenes and diverse, challenging human motions. The experimental result demonstrates the effectiveness of HSC4D. Our work contributes to extending the motion capture to large dynamic scenes. We hope this work will foster the creation and interaction of the human-dynamic digital world in the future.

{\small
\bibliographystyle{ieee_fullname}
\bibliography{egbib}
}

\end{document}